\pdfoutput=1
\documentclass{article}

\usepackage[style=numeric]{biblatex}
\usepackage[preprint,nonatbib]{nips_2018}

\usepackage[utf8]{inputenc} 
\usepackage[T1]{fontenc}    
\usepackage{hyperref}       
\usepackage{url}            
\usepackage{booktabs}       
\usepackage{amsfonts}       
\usepackage{nicefrac}       
\usepackage{microtype}      
\usepackage[pdftex]{graphicx}
\usepackage{subcaption}
\usepackage[group-separator={,}]{siunitx}

\title{Learning Joint Acoustic-Phonetic Word Embeddings}

\author{
  Mohamed El-Geish
  \\
  Voicea \\
  \texttt{geish@voicea.ai} \\
}

\begin{document}
\maketitle

\begin{abstract}
Most speech recognition tasks pertain to mapping words across
two modalities: acoustic and orthographic. In this work, we suggest
learning encoders that map variable-length --- acoustic or
phonetic --- sequences that represent words into fixed-dimensional
vectors in a shared latent space; such that the distance between
two word vectors represents how closely the two words sound.
Instead of directly learning the distances between word vectors,
we employ weak supervision and model a binary classification task
to predict whether two inputs, one of each modality, represent
the same word given a distance threshold. We explore various
deep-learning models, bimodal contrastive losses, and techniques for mining hard negative examples such as the semi-supervised technique of self-labeling. Our best model achieves an $F_1$ score of 0.95 for the binary classification task.
\end{abstract}

\section{Introduction}
The proliferation of voice-first applications and devices generated unprecedented demand to improve a plethora of speech recognition tasks. In this work, we propose a building block for speech recognition tasks like automatic speech recognition (ASR), keyword spotting, and query-by-example search. The outcome is an embedding model that maps variable-length --- acoustic or phonetic --- sequences that represent words into fixed-dimensional vectors in a shared latent space, which connects audio and phonetic modalities together, and encompasses a distributed representation of words, such that the distance between two word vectors represents how closely the two words sound: The more similar words sound, in either modality, the closer they end up in the shared vector space. Deep learning is befitting to learn pairwise relationships and joint embeddings, which have become the cornerstone of many machine learning applications~\cite{bromley1994signature, schroff2015facenet, harwath2017learning, li2018survey}. One of the applications we can reformulate is using the learned joint embeddings is ASR hypotheses reranking to reduce word error rate~\cite{bengio2014word, ma2017asr}.

The approach we took to learn distances between word embeddings in the vector space can be described as a weakly supervised task: Instead of training using ground-truth pairwise distances, we produced ground-truth data of word pairs labeled as either similar-sounding (distance of 0) or different-sounding (distance of 1) --- making it a binary classification problem. At inference time, the model predicts  real-valued distances that can be turned into labels given a distance threshold.

Despite the emergence of many successful embedding models, learning them is relatively poorly understood~\cite{histogramloss}. We experimented with various techniques to mine hard examples for training. Training using words picked at random from a corpus to be dissimilar to one another impedes learning; the model needs to be trained using hard examples of words that sound slightly different as examples of distant words~\cite{schroff2015facenet, wu2017sampling}. Another facet of weak supervision is creating ground-truth data using heuristics applied to the results of ASR systems and unreliable, non-expert transcribers.

\section{Related Work}
Thanks to recent natural language processing (NLP) research~\cite{mikolov2013word2vec, pennington2014glove, abdulkader2016deeptext}, the use of word embeddings to represent the semantic relationships between words has been prevailing. Similarly, in the field of speech recognition, the use of acoustic word embeddings has improved many tasks like keyword spotting~\cite{chen2015query}, ASR systems~\cite{bengio2014word}, query-by-example search~\cite{levin2013fixed, levin2015segmental, settle2017query}, and other word discrimination tasks~\cite{settle2016discriminative, kamper2016deep}; it also enabled attempts at unsupervised speech processing~\cite{kamper2017segmental, chung2016audio} and representing semantic relationships between words~\cite{chung2017learning, chung2018speech2vec}. The aforementioned speech recognition research learned models from the acoustic representation of words --- a single view, which requires side information to map that view into text. Infusing cross-modal (acoustic and orthographic) representations of words into the embedding model showed improvements in a plethora of applications; for example, \cite{he2016multi} jointly learned to embed acoustic sequences and their respective character sequences into a common vector space using deep bidirectional long short-term memory (LSTM) embedding models and multi-view contrastive losses. In~\cite{chung2018unsupervised}, the model separately learns acoustic and orthographic embedding spaces, and then attempts to align them using adversarial training; it employs a bidirectional LSTM encoder and a unidirectional LSTM decoder to perform the cross-modal alignment in an unsupervised fashion; the performance of such model is comparable to its supervised counterpart.

One of the limitations of mapping acoustic representations to their corresponding character sequences is confusing homographs (words that share the same spelling but not the same pronunciation) --- they may miss the subtle phonetic variations within words and/or across dialects. Multiple approaches attempted to mitigate such limitation. For example, \cite{lim2018learning} models phonetic information by training using both word- and frame- level criteria; however, it doesn't learn from phonetic labels so the learned embeddings cannot precisely represent the phonetic spelling of words. In~\cite{synnaeve2014phonetics}, the objective is to learn word embeddings from audio segments and use side information (phonetic labels) to train an acoustic model, which can be used for ASR systems --- a desideratum similar to ours; it trains the model using a Siamese neural network and explores multiple loss functions. In~\cite{chung2016audio}, the objective is to find vector representations that precisely describe represent the sequential phonetic structures of audio segments; in addition, \cite{he2016multi} considers directly training using phonetic-spelling supervision as a future direction. It's worth mentioning that we  discovered the latter after embarking on this endeavor, which validated our decision of using phonetic spelling instead of orthography.

In~\cite{bengio2014word, settle2016discriminative, he2016multi}, the models are trained using matched (similar-sounding) and mismatched (different-sounding) pairs of words; the mismatched words are drawn randomly, which makes minimizing losses like triplet and contrastive loss a challenge~\cite{wu2017sampling}. In~\cite{chung2018unsupervised}, adversarial training is used to align the acoustic and orthographic vector spaces. We use three different techniques, which we detail below, to mine for hard negative examples mining.

\section{Dataset and Representations}
The raw data consist of 25k short, single-channel recordings and their respective transcripts. The recordings were captured at a sample rate of 16kHz and encoded using pulse-code modulation with 16-bit precision. Transcripts do not include word alignments and may include errors. The dataset is proprietary and growing. To obtain the phonetic spelling of words, we used the LibriSpeech lexicon~\cite{librispeech}; it contains over 200k words and their pronunciations; out-of-vocabulary (OOV) words are excluded from the acoustic dataset as they lack ground-truth phonetic labels.

In order to obtain word alignments, we force-aligned each transcript in the raw dataset with recognition hypotheses generated by an ensemble of ASR systems. When the forced alignment failed due to an insertion or a deletion error, we considered the affected words too noisy to include. When the human-selected transcript for a word aligned successfully with an ASR hypothesis, we labeled the pair (the audio segment and its phonetic label) as similar-sounding (distance of 0). Using the set of positive pairs, we mined unique pairs of ASR hypotheses that substituted the human-selected transcript for the same audio segment and labeled them as hard negative examples (distance of 1). The restriction of starting from positive pairs reduced the overall number of examples; however, it drastically increased the quality of the labels as we minimized the number of false negatives.

We employed another technique to mine more of the hard negative examples: We started from the hard negative examples we had created above and computed their phonetic-edit distance (between the human-selected transcript and the ASR's hypothesis for the same audio segment); then, we picked ones that scored below a maximum distance threshold (0.7 was satisfactory) and grouped them by their respective human-selected transcripts; finally, we synthesized additional unique hard negative examples for ones that share the same human-selected transcript. The result of this process increased the number of negative examples from 73,845 to 439,679; it also helped balance the dataset, which included 393,623 positive examples. We removed stop words from the dataset as they are not interesting for downstream tasks; words shorter than 0.2 seconds long were also removed as manual inspection deemed them mostly mislabeled. The processed dataset consisted of 654,224 examples that were split, with class-label stratification, into train/dev/test datasets of sizes 621,905/12,692/19,627 respectively.The three sets contained 352,769/11,775/17,632 unique audio segments, respectively.

\begin{figure}[ht]
  \centering
  \begin{subfigure}{.5\textwidth}
    \centering
    \includegraphics[width=0.8\linewidth]{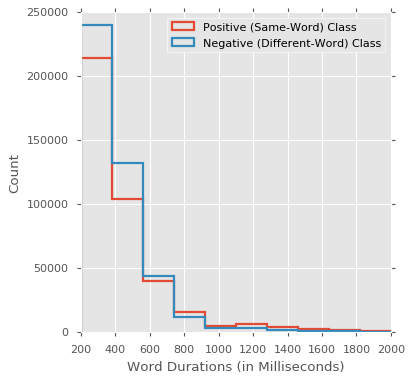}
  \end{subfigure}%
  \begin{subfigure}{.5\textwidth}
    \centering
    \includegraphics[width=0.8\linewidth]{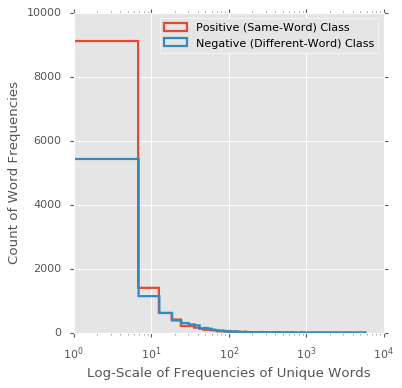}
  \end{subfigure}
  \caption{Distributions of word durations (left) and frequencies (right) in the dataset.}
\end{figure}

The third technique we used to mine even more hard negative examples is a semi-supervised learning technique that's attributed to Scudder for his work circa 1965: self-labeling~\cite{scudder1965probability}. Every $r$ epochs during training, we allow the dataset generator to augment the training dataset with newly minted hard negative examples. We encode the phonetic spelling of the training set's lexicon and build a $k$-dimensional tree for the resultant phonetic embeddings. Then we sample, at random, from the unique audio segments in the training set such that the sample's size is proportional to the number of epochs. Constraining the contributions of the less accurate models at early epochs reduces the chances of polluting the training set with easy-to-predict negative examples and hogging the memory. For each audio segment in the sample, we compute an embedding and find the closest phonetic neighbors within a maximum distance that reflects similarity (predicted a positive label). For each neighbor, we look up the true label of the word pair (audio segment and its phonetic label); if the model made a new mistake (the true label was negative and the pair is unique), we add the example to the training set as a hard negative. We observed a healthy growth in the training set's size thanks to self-labeling; it reached a 41\% relative growth (to a total of 876,771 training examples) at one point.

We represented audio as mel-spectograms using standard parameters given a sample rate of 16KHz: 400 samples (25ms) for the length of the Fast Fourier Transform (FFT) window; a hop of 200 samples between successive frames, and 64 mel bands. The input audio signal was centered and padded with zeros (silence) to fit in a window of 2 seconds to fix the size of the model's input; the result is a tensor of size $64\times161\times1$ (single channel). Audio examples were normalized individually using cepstral mean and variance normalization~\cite{strand2004cepstral} and then normalized across examples to have zero mean and unit variance for each mel band.

We represented phonetic spelling of words as a sequence of one-hot encoded vectors; the longest word in the LibriSpeech lexicon has 20 phones and there are 69 unique phones in the ARPAbet flavor used by the lexicon. We indicate an empty phone using a sentinel value (making the length of the alphabet 70). The result is a matrix of size $70\times20$.

\section{Methods}
In this work, we follow an approach similar to~\cite{he2016multi} in the sense that we model the task at hand as a weakly supervised, bimodal binary classification task; and along the way, we learn acoustic and phonetic encoders that map words into a shared vector space such that the distance between two word vectors represents how closely the two words sound. We train a Siamese neural network~\cite{bromley1994signature} that feeds forward the acoustic and phonetic representations of a pair of words through a series of transformations to encode the two inputs into $\ell^2$-normalized vectors (embeddings), then outputs the distance between the two embeddings. Our objective is to minimize the contrastive loss~\cite{hadsell2006dimensionality}, which allows us to learn encoders that map similar inputs to nearby points and dissimilar inputs to distant points in the shared vector space. To describe the objective formally, let $(x_a,\, x_p)$ be the input word pair (acoustic and phonetic representations, respectively) and $y$ be its true binary label such that $y = 0$ when the two representations, acoustic and phonetic, are of the same word; otherwise, $y = 1$. The model learns two functions, $f(x_a)$ and $g(x_p)$, that map the inputs into $\ell^2$-normalized embeddings; for each prediction, we compute the distance $\mathcal{D}(f(x_a),\, g(x_p))$, or $\mathcal{D}$ for brevity, between the outputs of both functions using a distance function such as the Euclidean or cosine distance. Since the embeddings are $\ell^2$-normalized, we think of the two distance functions as mostly interchangeable. Given $N$ training examples, we minimize the following function:
\[
\mathcal{L} = \frac{1}{N} \sum_i^N \big[ \, (1 - y^{(i)}) \, (\mathcal{D}^{(i)})^2 + y^{(i)} \, \max(0, \, m - \mathcal{D}^{(i)})^2 \, \big]
\]

where $m > 0$ is a margin parameter that controls when dissimilar pairs contribute to the loss function: only when $\mathcal{D} < m$. Unless otherwise specified, we used $m = 1$ for the experiments below.

\section{Experiments and Results}
Evaluating embeddings depends heavily on the downstream task; in this work, we picked $F_1$ score (the harmonic mean of precision and recall) as the metric of choice for the bimodal binary classification task. The binary labels at test time were calculated at a distance threshold of $0.5$, which approximates the observed break-even point when $m = 1$ in our experiments. In order to compute the break-even point for the test set, we generated all unique pairs of acoustic and phonetic inputs in the set --- approximately, 192.6 million word pairs. The $F_1$ score of our best-performing model is $0.95$.

\begin{figure}[ht]
  \centering
  \begin{subfigure}{.5\textwidth}
    \centering
    \includegraphics[width=0.8\linewidth]{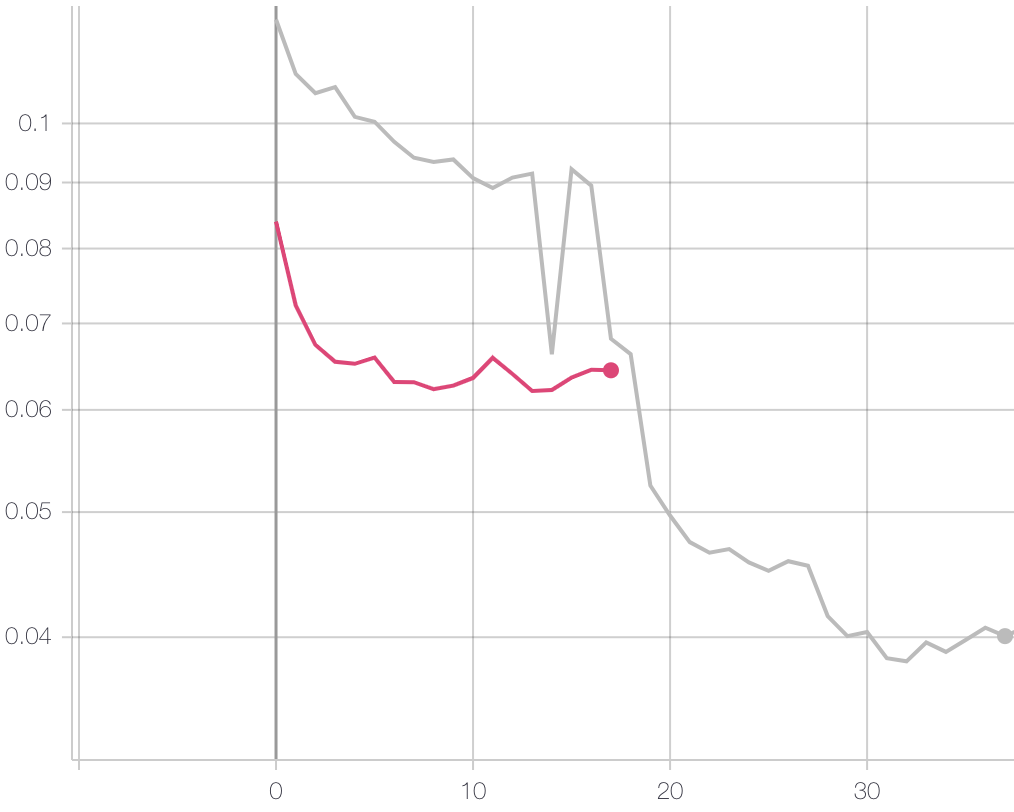}
    \caption{Development set loss before self-labeling to augment training data (red line) and after (gray line) — learning use to stagnate at much earlier epochs (x-axis).}
  \end{subfigure}%
  \begin{subfigure}{.5\textwidth}
    \centering
    \includegraphics[width=0.8\linewidth]{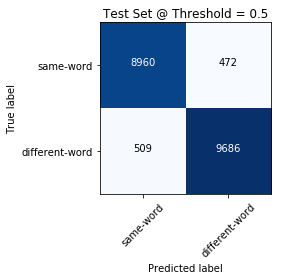}
    \caption{Confusion matrix of our best-performing model.}
  \end{subfigure}
\end{figure}

Projecting a sample of embeddings derived from the test set using a t-SNE~\cite{maaten2008visualizing} model results in a reasonable clustering of words. More interestingly, the model found phonetic analogies such as "cat" to "cool" is like "pat" to "pool", etc. We also inspected a sample of 50 classification mistakes the model made. A few patterns emerged: the model failed to predict a distance that reflects similarity between the acoustic and phonetic embeddings when the audio was too noisy, too faint (far field), or the speech was accented. Cross talk and reverb were also problematic but not as common. Audio preprocessing to clean up the signal may be helpful in such cases. Also, training using n-grams and acquiring more data from speakers with accents and different acoustic environment may boost performance as we expect distributions of words to be non-IID (independent and identically distributed).

We experimented with and manually tuned various hyperparameter choices for input representations, architecture, contrastive losses, batch size, number of hidden units, number of epochs, embedding size, dimensions of filters in convolutional neural networks (CNN), single vs. bidirectional LSTM cells, Euclidean- vs. cosine- based distance functions, etc. The sparse representation of phonetic input led us to believe that the architecture for its encoder should be different from that of the acoustic one; however, empirical evidence suggests that mirroring the same architecture for both encoders, with the exception of a dropout for the acoustic input layer, yields better results. Table~\ref{table:experiments} summarizes notable experiments and their results.

\renewcommand{\arraystretch}{1.2}
\begin{table}[ht]
\caption{Summary of notable experiments and their results for training and testing, respectively.}
\label{table:experiments}
\centering
\begin{tabular}{c l c} 
 \toprule
 \textbf{\#} & \textbf{Notable Experiment Details} & \textbf{$F_1$ Scores} \\ [0.5ex]
 \toprule
 1  & CNN ($3\times3\times32$); dense layer; 256-D embedding; batch size = $32$ & $0.97, 0.91$ \\ \hline
 2  & CNN ($3\times3\times32$) -> ($3\times3\times64$); dense layer; $256$-D embedding;\\
    & batch size = $32$ & $0.99, 0.93$ \\ \hline
 3  & Same as \#2 but for a dropout with a rate of $0.5$ after the first hidden layer & $0.99, 0.94$ \\ \hline
 4  & Same as \#3 but with another dropout of $0.5$ after the second hidden layer & $0.95, 0.91$ \\ \hline
 5  & Same as \#3 but with margin = the phonetic-edit distance for the pair & $0.96, 0.91$ \\ \hline
 6  & Same as \#3 but with incoming weights constrained to a maximum norm of $3$ & $0.99, 0.94$ \\ \hline
 7  & 2 unidirectional LSTM layers with $128$ hidden units; dense layer; \\
    & $256$-D embedding; batch size = $32$; $24$ epochs (in $99$ hours) & $0.91, 0.87$ \\ \hline
 8  & 2 Bidirectional LSTM layers with $512$ hidden units and a dropout of $0.4$\\
    & in between; a dropout of $0.2$ for the acoustic input; $512$-D embedding;\\
    & $28$ epochs (in $47$ hours) & $0.95, 0.91$ \\ \hline
 9  & CNN with 2 blocks [($3\times3\times64$) -> ($2\times2$) max pooling]; two dense\\
    & layers with 512 hidden units and a dropout of 0.4 in between; a dropout\\
    & of 0.2 for the acoustic input; $512$-D embedding; cosine distance;\\
    & batch size = $128$; 64 epochs (in 4.8 hours) & $0.99, \textbf{0.95}$ \\ \hline
 10 & Same as \#9 but with additional dropout of $0.4$ between convolutional\\
    & layers as well; trained for much longer (142 epochs in 19 hours) & $0.96, 0.93$ \\
 \bottomrule
\end{tabular}
\vspace{-5mm} 
\end{table}

We use the $\ell^2$-normalized output of each encoder's last layer as the learned embedding hence we don't use an activation function to those layers. Unless otherwise specified, we constrained incoming weights to a maximum norm of $3$ for layers with dropout regularization to allow for large learning rates without the risk of the weights ballooning~\cite{srivastava2014dropout}. We used the Adam optimizer~\cite{kingma2014adam} with initial learning rates tuned for different architectures ($0.001$ for CNN and $0.0001$ for LSTM); we reduced the learning rate by a factor of $2$ when learning stagnates for $2$ epochs. Since the addition of self-labeling to our experiments, we observed a much steeper acceleration in learning in early epochs. To balance the learning rate decay with the addition of new data on-the-fly during training, we increase the learning rate --- by the same factor to a maximum of $0.001$ --- when new hard negative examples are mined to give the model a better chance at learning from the newly minted examples.

For LSTM, we used $tanh$ for activation; otherwise, we used rectified linear units (ReLU)~\cite{nair2010rectified}, with the exception of the model's and the encoder's output layers. Network weight were initialized using He initialization~\cite{he2015delving} when ReLU was used; otherwise, we use the Xavier initialization method~\cite{glorot2010understanding}.

\section{Conclusion and Future Work}
 We presented techniques for learning functions that map acoustic and phonetic representations of words into fixed-dimensional vectors in a shared latent space, which are extremely useful in a plethora of speech recognition tasks. We experimented with many modeling techniques, hyperparameters, and neural network architectures to learn the joint embeddings; our best model is a Siamese CNN that feeds forward acoustic and phonetic inputs and brings together, in the shared vector space, similar-sounding words while keeping apart different-sounding ones. We used binary classification as a surrogate task to learn the embeddings at the last layer of each encoder. The choice of training examples cannot be cannot be overstated: we use three different techniques, including self-labeling, to mine hard negative examples for the contrastive loss function so that it can learn the subtleties requisite to discriminate between input words.
 
One of the areas to explore in future work is training multiple models for multiple word duration buckets to minimize extraneous padding. We'd also like to explore other loss functions detailed in~\cite{histogramloss, he2016multi}.
 
\subsubsection*{Acknowledgments}
We'd like to thank Ahmad Abdulkader for his mentorship and guidance; and the authors of related work and software tools~\cite{chollet2015keras, scikit-learn, mckinney2010data, hunter2007matplotlib, kluyver2016jupyter, jones2014scipy, mcfee2019librosa, paszke2017pytortch, nltk, sox, kaldi, ffmpeg} for allowing us to build on top of what they've created.

\printbibliography

\end{document}